\documentclass[10pt,twocolumn,letterpaper]{article}
\pdfoutput=1 

\usepackage{iccv}
\usepackage{times}
\usepackage{epsfig}
\usepackage{graphicx}
\usepackage{amsmath}
\usepackage{amssymb}

\usepackage{caption}
\usepackage{subcaption}

\usepackage[breaklinks=true,bookmarks=false]{hyperref}

\usepackage[accsupp]{axessibility}  % Improves PDF readability for those with disabilities.

\iccvfinalcopy % *** Uncomment this line for the final submission

\def\iccvPaperID{10244} % *** Enter the ICCV Paper ID here

\ificcvfinal\fi

\newcommand{\Vector}[1]{\lowercase{\mathbf{#1}}}
\newcommand{\Matrix}[1]{\uppercase{\mathsf{#1}}}

\newcommand{\PAR}[1]{\vskip4pt \noindent{\bf #1~}}

\begin{document}

%%%%%%%%% TITLE
\title{P1AC: Revisiting Absolute Pose From a Single Affine Correspondence}

\author{Jonathan Ventura$^1$ \quad Zuzana Kukelova$^2$ \quad Torsten Sattler$^3$ \quad D\'{a}niel Bar\'{a}th$^4$\\
$^1$ Department of Computer Science \& Software Engineering, Cal Poly, San Luis Obispo\\ 
$^2$ Visual Recognition Group, Faculty of Electrical Engineering, Czech Technical University in Prague\\  
$^3$ Czech Institute of Informatics, Robotics and Cybernetics, Czech Technical University in Prague\\
$^4$ Computer Vision and Geometry Group, ETH Z\"{u}rich
}

\maketitle
\ificcvfinal\thispagestyle{empty}\fi

%%%%%%%%% ABSTRACT
\begin{abstract}
Affine correspondences have traditionally been used to improve feature matching over wide baselines. While recent work has successfully used affine correspondences to solve various relative camera pose estimation problems, less attention has been given to their use in absolute pose estimation.   We introduce the first general solution to the problem of estimating the pose of a calibrated camera given a single observation of an oriented point and an affine correspondence.  The advantage of our approach (P1AC) is that it requires only a single correspondence,  in comparison to the traditional point-based approach (P3P), significantly reducing the combinatorics in robust estimation.  P1AC provides a general solution that removes restrictive assumptions made in prior work and  is applicable to large-scale image-based localization.  We propose a minimal solution to the P1AC problem and evaluate our novel solver on synthetic data, showing its numerical stability and performance under various types of noise.  On standard image-based localization benchmarks we show that P1AC achieves more accurate results than the widely used P3P algorithm. Code for our method is available at \url{https://github.com/jonathanventura/P1AC/}.
\end{abstract}

%%%%%%%%% BODY TEXT
\section{Introduction}
\label{sec:introduction}

\begin{figure}[t]
    \centering
    \includegraphics[width=2.4in]{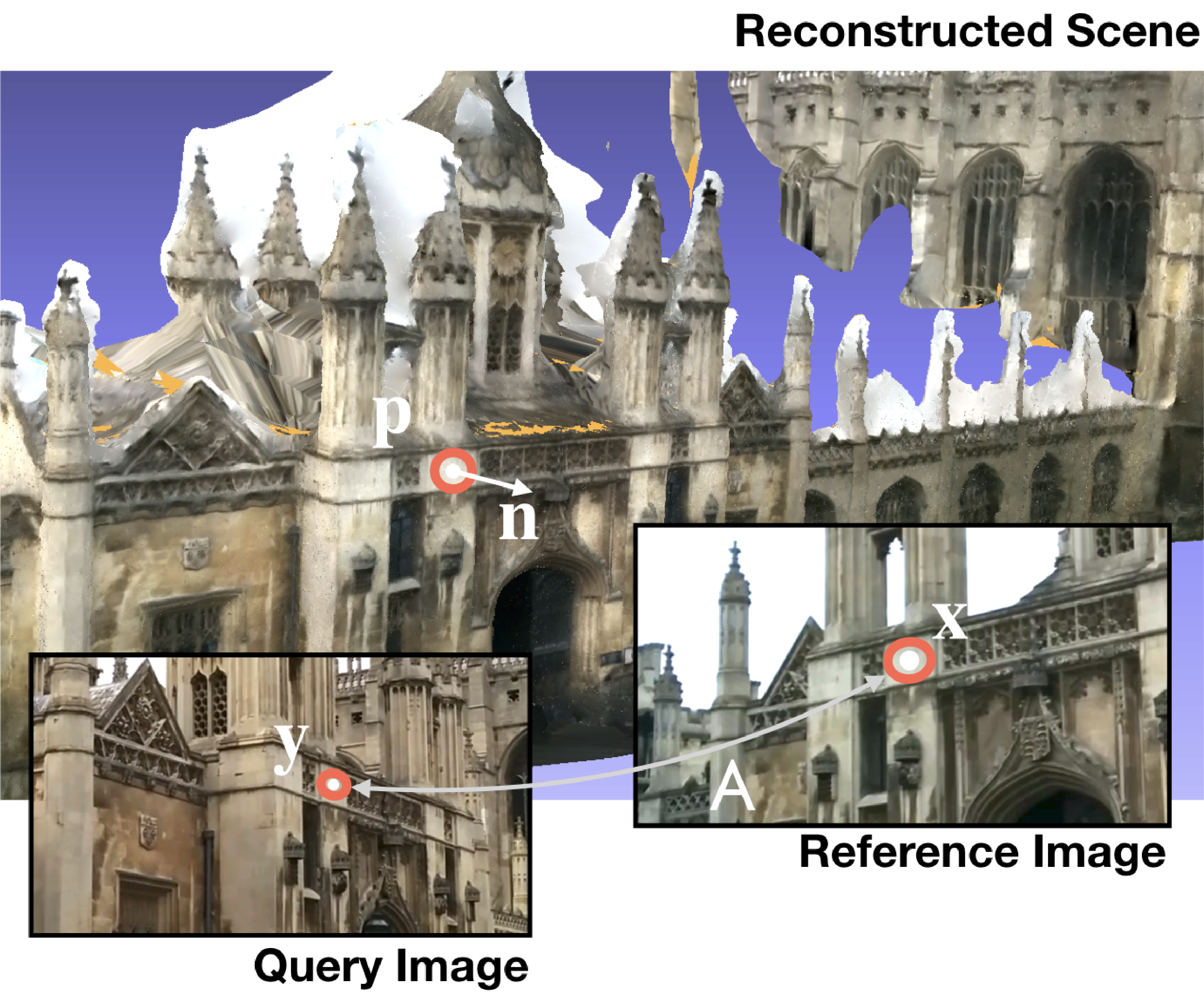}
    \caption{\label{fig:intro}Overview of the  perspective-one-affine-correspondence (P1AC) problem.  Our goal is to estimate the pose of a query image based on a single 2D observation of a known oriented 3D point and an affine correspondence to a reference image with known pose.}
\end{figure}

Image-based localization is the process of determining the pose of a query image in a reference coordinate system from pre-registered images in a database.  Image-based localization is an important technology for many applications including robotics \cite{mur2017orb}, self-driving cars \cite{hane20173d} and augmented reality \cite{ventura2014global}.  It also forms an essential component of large-scale structure-from-motion pipelines \cite{schoenberger2016sfm}.

The dominant method for image-based localization is to estimate the camera pose from correspondences between 2D features in the query image and known 3D points in the scene.  Since these correspondences are likely to contain mismatches, a best-fit model is found through Random Sample Consensus (RANSAC) \cite{fischler1981random} or one of its derivatives \cite{chum2003locally,Lebeda2012BMVC,barath2021graph}, where camera pose hypotheses are estimated from minimal samples of three point correspondences (PCs) using a perspective-three-point (P3P) algorithm \cite{persson2018lambda}.

In cases of very wide-baseline matching or with extreme differences in viewing angle, affine-covariant features increase match reliability in comparison to traditional difference-of-Gaussian detectors such as SIFT \cite{lowe2004distinctive}.  A match between affine-covariant features, which we will call an affine correspondence (AC), estimates a $2\times2$ affine transformation matrix between the local regions around the corresponding points in the two images.

Recently, several authors have developed a theory of the constraints induced by ACs on two-view geometry \cite{bentolila2014conic,raposo2016theory,eichhardt2018affine}, and minimal solvers for relative pose estimation from ACs \cite{eichhardt2018affine,eichhardt2020relative}.  
However, relatively little attention has been given to absolute pose estimation from ACs.  We revisit the problem of estimating absolute camera pose from a single affine correspondence, and refer to this as the perspective-one-affine-correspondence (P1AC) problem.  K\"{o}ser and Koch \cite{koser2008differential} considered a limited form of the P1AC problem by assuming that the reference view is orthorectified.  In contrast, our method removes this limiting assumption and makes more general assumptions: that the reference and query images are perspective images of a locally planar surface in an arbitrary orientation, that we have estimated the normal to the surface at the observed 3D point, and that we have estimated an affine transformation between the reference and the query images.  We derive a system of equations for this problem and propose the first general solution for absolute pose from a single AC. 

In comparison to P3P solutions, our P1AC solution only requires a single correspondence instead of three in the minimal case, and thus significantly reduces the complexity of robust estimation.   We demonstrate these advantages through experiments on synthetic and real data.

\noindent
Our contributions are as follows:
\begin{itemize}
\vspace{-1.6ex}
\item We introduce a novel solution to the P1AC problem using reduction to the 3Q3 problem \cite{kukelova2016efficient}.
Our solution is general and does not impose the restrictive assumptions made in previous work.
\vspace{-1.6ex}
\item Experiments on synthetic data establish the numerical stability of our solver and analyze its performance under various types of noise.
\vspace{-1.6ex}
\item Experiments on two benchmark datasets for image-based localization show that our P1AC solution produces more accurate pose estimates than P3P. 
\end{itemize}

\section{Related work}
\label{sec:related_work}

Affine correspondences are commonly estimated using either frame-to-frame tracking techniques \cite{lucas1981iterative,baker2004lucas} or affine-covariant region detection \cite{mikolajczyk2004scale, mishkin2018repeatability}.  In the case of frame-to-frame tracking, the deformation of a locally planar surface observed by a moving camera is well-modeled by an affine transformation.  For affine-covariant regions, an affine transformation describing the shape of the local region around the interest point is estimated \cite{baumberg2000reliable,mishkin2018repeatability}.  
While computing ACs may increase feature detection computation time over computing PCs, ACs improve match reliability in wide-baseline image matching \cite{IMC2020}, introduce new constraints on camera geometry \cite{koeser2009,bentolila2014conic} and, therefore, decrease the number of correspondences needed for camera pose estimation \cite{koser2008differential,raposo2016theory,eichhardt2018affine,eichhardt2020relative}.

Recent work has examined how ACs might be used to improve and simplify pose estimation and subsequently improve image-based localization, visual odometry, and structure-from-motion.  K\"{o}ser developed a theory of relationship between homographies and ACs \cite{koeser2009}, and Bentolila and Francos demonstrated that an AC provides three constraints on the fundamental matrix \cite{bentolila2014conic}.  Raposo and Barreto re-derived these constraints \cite{raposo2016theory}, making them usable for essential matrix estimation using only two ACs instead of five PCs as is normally needed \cite{nister2004efficient}.  Eichhardt and Chetverikov extended relative pose estimation using ACs to handle any central camera model \cite{eichhardt2018affine}.  
Eichhardt and Barath \cite{eichhardt2020relative} demonstrated relative pose estimation from a single AC and a depth map estimated by a neural network.
Recently, Barath et al.~provided recommendations for speeding up RANSAC when using ACs \cite{barath2020making}, including local feature geometry refinement and uncertainty propagation in RANSAC. 
The ideas presented in \cite{barath2020making} can be used in combination with any pose estimation solver that uses ACs, including the solvers proposed in this paper.

Absolute pose estimation using ACs has received less attention than relative pose.
Algorithms for absolute pose estimation from PCs have a long history, dating back to Grunert's solution 
\cite{grunert1841das}.  Among many recent P3P solutions \cite{kneip2011novel,ke2017efficient,persson2018lambda,ding2023revisiting} the Lambda Twist method~\cite{persson2018lambda} and~\cite{ding2023revisiting}  are among the fastest and most numerically stable. 

K\"{o}ser and Koch introduced a solution for absolute pose from a single AC called Differential Pose Resection (DPR), with the restriction that the reference image is a scaled orthographic image of a planar surface with the normal parallel to the optical axis \cite{koser2008differential}.  In our setting, the reference image is, instead, a perspective image of a locally planar surface in an arbitrary orientation. Their method could be applied in our setting if we rectified the image and then re-computed the affine region in the rectified image. 
However, this is expensive as it needs to be done per feature.  An alternative is to virtually warp the AC so that the reference view is orthorectified.  
However, this warping cannot be done exactly, and, as we show in our experiments, this approximate warping leads to decreasing accuracy in the DPR solution as the amount of perspective distortion increases. 

Collins and Bartoli \cite{collins2014infinitesimal} derived novel constraints and a minimal solution for the absolute pose of a camera given four co-planar PCs.  They first calculate the homography between the object plane and the camera image. 
 While the homography could be directly decomposed into the camera pose, they propose an alternative solution which addresses the spatially-varying accuracy of the homography.  They choose an optimal point on the plane to compute a first-order approximation to the homography (an affine transformation) and then compute the pose of the camera based on the 2D point projection and the first-order approximated local transformation.  Their solution, called Infinitesimal Plane-based Pose Estimation (IPPE), is therefore highly related to DPR \cite{koser2008differential}, in that it ultimately computes the pose of the camera based on a single AC between the object plane and the image.   To apply IPPE in our setting, we generate three virtual PCs around the AC in the reference image by ray intersection with the object plane.  However, as we show in our experiments, the IPPE solution is less accurate than ours because of the inherent inaccuracy in the virtual PCs.

Haug and J\"{a}hne developed a solution for computing the absolute pose of a stereo or RGB-D camera from a single AC \cite{haug20106}.  Their solution assumes knowledge of the depth of the point in both the reference and query images and thus is not applicable in our  scenario of localizing a monocular camera, where the depth of the point in the query image is unknown.

\section{Minimal solvers}
\label{sec:minimal_solver}

\paragraph{Notation:} We use a sans-serif capital letter $\Matrix{M}$ for a matrix, a bold lower-case letter $\Vector{v}$ for a vector, and an italic lower-case letter $s$ for a scalar.  We use subscripts to indicate matrix and vector indexing.
A tilde over a vector indicates a homogeneous version of that vector ($\tilde{\Vector{x}}=[~x_1~x_2~1~]^\text{T})$.  

Consider two calibrated pinhole cameras observing a scene: a reference camera, whose pose w.r.t.\ the world coordinate system is known, and a query camera, whose pose is unknown.
Without loss of generality, we assume that the reference camera has extrinsics $[~\Matrix{I}~|~\Vector{0}~]$, and the query camera has extrinsics $[~\Matrix{R}~|~\Vector{t}~]$.\footnote{%In general, w
When the reference camera is not at the origin, we first transform the world coordinate system to place the reference camera at the origin and then apply the inverse transform to the resulting solution for $\Matrix{R}$ and $\Vector{t}$.} 
Here $\Matrix{R}$ and $\Vector{t}$ represent the world-to-camera transformation of the query camera, 
transforming points from  the world coordinate system to the coordinate system of the query camera. 
This is a standard setup in localization pipelines, where our goal is to estimate the 3D rotation $\Matrix{R}$ and translation $\Vector{t}$ of the query camera.  

The inputs to our P1AC method are: corresponding points $\Vector{x}$ and $\Vector{y}$ in the reference and query images, respectively; $\Matrix{A}$, the affine transformation from the local region around $\Vector{x}$ to the local region around $\Vector{y}$; $d$, the depth of the observed 3D point in the reference image; and $\Vector{n}$, the surface normal at the 3D point in the coordinate system of the reference image.

First, we will develop a system of constraints on the absolute pose of the query camera induced by the AC, and then we will describe a minimal solution to this derived system.  

\subsection{Constraints induced by AC on camera pose}

Let us define a function $f:\mathbb{R}^2\rightarrow\mathbb{R}^2$ that maps points in the reference image to points in the query image. Let $\Vector{x},\Vector{y}$ be corresponding points in the two images such that $\Vector{y}=f(\Vector{x})$, and let $\Vector{u}$ be a point in the neighborhood of $\Vector{x}$.  
The first-order Taylor approximation of $f$ centered at $\Vector{x}$ is $\hat{f}(\Vector{u})=\Vector{y}+\Matrix{A}(\Vector{u}-\Vector{x})$, where $\Matrix{A}=\nabla_{\Vector{u}} f(\Vector{u})|_{\Vector{u}=\Vector{x}}$ is the local affine transformation matrix.  

We now aim to analytically derive the form of $\nabla_{\Vector{u}} f(\Vector{u})$ under perspective projection.  Let $\pi^{-1}:\mathbb{R}^2\rightarrow\mathbb{R}^3$ be the mapping from a point in the reference image to the local 3D surface.   Then we have \begin{equation}
    f(\Vector{u})=\pi(\Matrix{R}\pi^{-1}(\Vector{u})+\Vector{t}).
\end{equation}

The projection function $\pi(\Vector{p})$ is straightforward: $\pi(\Vector{p})=[~p_1/p_3,~p_2/p_3~]^\text{T}$. The un-projection function $\pi^{-1}(\Vector{u})$, however, requires knowledge of the scene geometry.  
We assume that the cameras observe a locally planar surface defined by 3D point $\Vector{p}=d \tilde{\Vector{x}}$ and normal $\Vector{n}$.To find $\pi^{-1}(\Vector{u})$, we intersect ray $\Vector{u}$ with the plane. The plane is defined by
\begin{equation}
    \Vector{n}^\text{T} (\Vector{p}' - \Vector{p})=0
\end{equation}
where $\Vector{p}'$ is a point on the planar surface.  Parameterizing the ray-plane intersection by $\Vector{p}'=\alpha \tilde{\Vector{u}}$, we have
\begin{equation}
    \Vector{n}^\text{T} (\alpha \tilde{\Vector{u}}-\Vector{p})=0
\end{equation}
and therefore
\begin{equation}
    \alpha = 
    \frac{\Vector{n}^\text{T} \Vector{p}}
    {\Vector{n}^\text{T} \tilde{\Vector{u}}}.
\end{equation}
Thus, under the assumption of a locally planar surface, we have determined $\pi^{-1}(\Vector{u})$ as follows:
\begin{equation}
\label{eq:unproject}
    \pi^{-1}(\Vector{u})=\frac{\Vector{n}^\text{T} \Vector{p}}
    {\Vector{n}^\text{T} \tilde{\Vector{u}}}
    \Vector{\tilde{u}}.
\end{equation}

Let $\Vector{q}=\Matrix{R}\pi^{-1}(\Vector{u})+\Vector{t}$ so that $f(\Vector{u})=\pi(\Vector{q})$.  To compute $\nabla_{\Vector{u}} f(\Vector{u})$, we first apply the quotient rule:
\begin{align}
\label{eq:Jf}
\nabla_{\Vector{u}} f(\Vector{u})  &= \nabla_{\Vector{u}} \pi(\Vector{q})\nonumber 
%\\
= \nabla_{\Vector{u}} \left( \frac{1}{q_3} \begin{bmatrix} q_1 \\ q_2 \end{bmatrix} \right)\nonumber \\
&=
\frac{1}{q_3^2} \left[
q_3 \left( \nabla_{\Vector{u}} \begin{bmatrix} q_1 \\ q_2 \end{bmatrix} \right) -
\begin{bmatrix}q_1 \\q_2 \end{bmatrix} (\nabla_{\Vector{u}} q_3)
\right]\nonumber \\
& =
\frac{1}{q_3} \left[
\nabla_{\Vector{u}} \begin{bmatrix}q_1\\q_2 \end{bmatrix} - \frac{1}{q_3}\begin{bmatrix}q_1\\q_2\end{bmatrix} (\nabla_{\Vector{u}} q_3) 
\right] \nonumber \\
& =
\frac{1}{q_3} \left[
\nabla_{\Vector{u}} \begin{bmatrix}q_1\\q_2 \end{bmatrix} - f(\mathbf{u}) (\nabla_{\Vector{u}} q_3) 
\right] .
\end{align}

Noting that $\Vector{q} = \alpha (\Matrix{R} \tilde{\Vector{u}})+\Vector{t}$,
to compute $\nabla_{\Vector{u}}\Vector{q}$ we apply the product rule:
\begin{equation}
\label{eq:Jzq}
\nabla_{\Vector{u}}\Vector{q} = 
(\Matrix{R} \tilde{\Vector{u}}) \nabla_{\Vector{u}}\alpha +
\alpha \nabla_{\Vector{u}}(\Matrix{R} \tilde{\Vector{u}}).
\end{equation}
To compute $\nabla_{\Vector{u}}\alpha$ we apply the quotient rule:
\begin{align}
\label{eq:Jualpha}
\nabla_{\Vector{u}}\alpha &= \nabla_{\Vector{u}} \left( \frac{\Vector{n}^T\Vector{p}}{\Vector{n}^T\tilde{\Vector{u}}} \right) 
= -\frac{(\Vector{n}^T\Vector{p})(\Vector{n}_{1:2}^T)}{(\Vector{n}^T\tilde{\Vector{u}})^2}
= -\alpha \frac{\Vector{n}_{1:2}^T}{\Vector{n}^T\tilde{\Vector{u}}} .
\end{align}
Note also that $\nabla_{\Vector{u}}(\Matrix{R} \tilde{\Vector{u}}) = \Matrix{R}_{:,1:2}$.  Now we plug our expressions for $\nabla_{\Vector{u}}\alpha$ (Eq. \ref{eq:Jualpha}) and $ \nabla_{\Vector{u}}(\Matrix{R} \tilde{\Vector{u}})$ into Eq.~\ref{eq:Jzq} to obtain
\begin{align}
\nabla_{\Vector{u}}\Vector{q} &= 
(\Matrix{R} \tilde{\Vector{u}}) (-\alpha \frac{\Vector{n}_{1:2}^T}{\Vector{n}^T\tilde{\Vector{u}}}) +
\alpha \Matrix{R}_{:,1:2}\\
&= \alpha\left( 
\Matrix{R}_{:,1:2} -
(\Matrix{R} \tilde{\Vector{u}}) \frac{\Vector{n}_{1:2}^T}{\Vector{n}^T\tilde{\Vector{u}}}
 \right) \\
 &= \frac{\Vector{n}^T\Vector{p}}{\Vector{n}^T\tilde{\Vector{u}}}
 \left( 
\Matrix{R}_{:,1:2} -
(\Matrix{R} \tilde{\Vector{u}}) \frac{\Vector{n}_{1:2}^T}{\Vector{n}^T\tilde{\Vector{u}}}
 \right).
\end{align}
Evaluating this expression at $\Vector{u}=\Vector{x}$ we obtain
\begin{equation}
\label{eq:Jq}
\nabla_{\Vector{u}} \Vector{q} |_{\Vector{u}=\Vector{x}}
=
d
 \left( 
\Matrix{R}_{:,1:2} -
(\Matrix{R} \tilde{\Vector{x}}) \frac{\Vector{n}_{1:2}^T}{\Vector{n}^T\tilde{\Vector{x}}}
 \right).
\end{equation}
Finally, we plug Eq.~\ref{eq:Jq} into Eq.~\ref{eq:Jf} to arrive at a formula for $\Matrix{J}=\nabla_{\Vector{u}} f(\Vector{u}) |_{\Vector{u}=\Vector{x}}$:
\begin{equation}
\label{eq:J}
\begin{split}
\Matrix{J}=\frac{d}{m}
(
\Matrix{R}_{1:2,1:2}({\Vector{n}^T\tilde{\Vector{x}}}) -
(\Matrix{R}_{1:2,:} \tilde{\Vector{x}}) \Vector{n}_{1:2}^T
 - \\
\Vector{y}
(
\Matrix{R}_{3,1:2}(\Vector{n}^T\tilde{\Vector{x}}) -
 (\Matrix{R}_{3,:} \tilde{\Vector{x}}) \Vector{n}_{1:2}^T
)
)
\end{split}
\end{equation}
where $m=\Vector{n}^\text{T} \tilde{\Vector{x}} (d (\Matrix{R}_{3,:} \tilde{\Vector{x}})+t_3)$.

We multiply $m$ on both sides of the constraint $\Matrix{A}=\Matrix{J}$ to cancel out the denominator and make the constraint linear.  Our set of six constraints to be solved is thus
\begin{align}
y_1 (\Matrix{R}_{3,:} \Vector{p} + t_3) - (\Matrix{R}_{1,:} \Vector{p} + t_1) = 0, \label{eq:1} \\
y_2 (\Matrix{R}_{3,:} \Vector{p} + t_3) - (\Matrix{R}_{2,:} \Vector{p} + t_2) = 0,\\
m a_{11} - m j_{11} = 0, \quad
m a_{12} - m j_{12} = 0, \\
m a_{21} - m j_{21} = 0, \quad
m a_{22} - m j_{22} = 0, \label{eq:6} 
\end{align}
where the first two constraints enforce that the 3D point projects to the 2D observation in the query camera, and the remaining constraints enforce that the affine transformation matrix equals the Jacobian matrix derived above.  Note that the only unknowns in equations \ref{eq:J} and \ref{eq:1}-\ref{eq:6} are elements of $\Matrix{R}$ and $\Vector{t}$.

\subsection{Minimal solution}

We have six independent constraints on the entries of $\Matrix{R}$ and $\Vector{t}$ and six degrees of freedom (three for the rotation and three for the translation).  Therefore, we can solve for the absolute pose with a single AC.

We use the Cayley parameterization of the rotation with parameters $x,y,z$ as follows:
\begin{equation}
    \label{eq:cayley}
    \Matrix{R} = \frac{1}{s}
    \begin{bmatrix}
     {\scriptstyle 1+x^2-y^2-z^2} & {\scriptstyle 2(xy-z)} & {\scriptstyle 2(y+xz)} \\ 
     {\scriptstyle 2(xy+z)} & {\scriptstyle 1-x^2+y^2-z^2} & {\scriptstyle 2(yz-x)} \\
     {\scriptstyle 2(xz-y)} & {\scriptstyle 2(x+yz)} & {\scriptstyle 1-x^2-y^2+z^2}
    \end{bmatrix}
\end{equation}
where $s=1+x^2+y^2+z^2$.   
Note that we cannot represent $180^\circ$ rotations with this parameterization; however, 
this degenerate configuration is not a problem in practice 
since it would mean that the reference and the query camera have no visual overlap.
After plugging this parameterization of $\Matrix{R}$ into equations \ref{eq:1}--\ref{eq:6}, we multiply all equations by $s$ to cancel out the denominator in $\Matrix{R}$.

Let $\Vector{c}=[st_1, st_2, st_3, x^2, xy, xz, y^2, yz, z^2, x, y, z, 1]^\text{T}$.
Writing the equations \ref{eq:1}-\ref{eq:6} as a matrix-vector multiplication $\Matrix{M}\Vector{c}=0$, we first eliminate $s\Vector{t}$ using Gauss-Jordan elimination.  This leaves three quadratic equations 
in ten monomials on 
$x,y,z$ -- a 3Q3 problem.  Kukelova et al.~\cite{kukelova2016efficient} introduced a fast solver for the 3Q3 problem which produces up to eight solutions for $x,y,z$.  
Once we obtain solutions for $x,y,z$, we solve for $\Vector{t}$ by backsubstitution into equations 
\ref{eq:1}-\ref{eq:6}.

\subsection{Alternative solvers}

We also consider two alternative solvers based on previous work, to establish a baseline comparison for the P1AC method proposed in this paper.

\subsubsection{Differential pose resection}

The Differential Pose Resection (DPR) method \cite{koser2008differential} solves the P1AC problem under the restriction that the reference image is an ortho-rectified view of the plane.  To apply the DPR method in the general case, we need to virtually warp the AC so that it comes from an ortho-rectified view.  

Let $\Vector{w}$ represent the 2D coordinates of a point on the plane and let $\Matrix{H}$ be a homography from the plane to the reference image s.t.~$\Vector{u} = \pi(\Matrix{H} \tilde{\Vector{w}})$
and
$\Vector{x} = \pi(\Matrix{H} [0 ~ 0 ~ 1]^\text{T})$.  
The point corresponding to $\Vector{u}$ is $\Vector{v} = g(\Vector{w})=f(\pi(\Matrix{H} \tilde{\Vector{w}}))$.  $g$ is thus a non-linear transformation from the plane to the query.  We replace $f$ with $\hat{f}$ since we only know the AC between the reference and query images.
We can approximate $g$ as an affine transformation by computing the matrix $\Matrix{A'}=\nabla_{\Vector{w}}g|_{\Vector{0}}$:
\begin{equation}
A' =  
\frac{A}{{H_{33}}}
\begin{bmatrix}
{\scriptstyle H_{11} - (H_{13}H_{31})/H_{33}} & {\scriptstyle H_{12} - (H_{13}H_{32})/H_{33}} \\
{\scriptstyle H_{21} - (H_{23}H_{31})/H_{33}} & {\scriptstyle H_{22} - (H_{23}H_{32})/H_{33}}
\end{bmatrix} \enspace .
\end{equation}

\subsubsection{Infinitesimal plane-based pose estimation}

The Infinitesimal Plane-based Pose Estimation (IPPE) method \cite{collins2014infinitesimal} solves for the pose of the query camera given four coplanar PCs.  In our case, the AC only contains a single PC.  To transform a P1AC problem into an IPPE problem, we generate three extra virtual PCs by sampling 2D points around $\Vector{x}$ in the reference image: the 3D points are found by ray intersection with the plane (Eq. \ref{eq:unproject}), and the 2D points are found by applying the affine matrix $\Matrix{A}$.  We generate three extra virtual PCs in this manner and provide them along with the PC from the original AC as input to IPPE.
It is important to note that these virtual correspondences are only approximations generated by the assumption that the AC lies on an actual plane, and that the AC is valid beyond an infinitesimal region around the 2D point. 

\section{Evaluation}
\begin{figure*}
    \centering
    \begin{subfigure}[b]{0.25\textwidth}
    \centering
    
    \includegraphics[width=\textwidth, trim=4mm 0mm 3mm 0mm, clip]{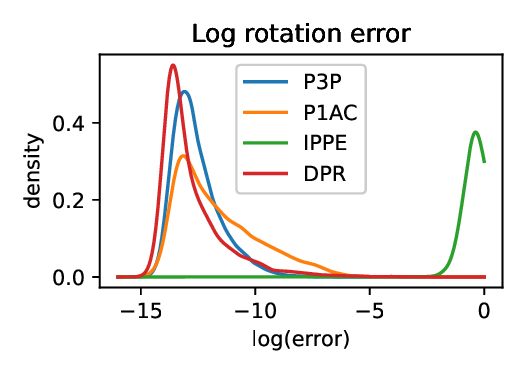}\\  \includegraphics[width=\textwidth, trim=4mm 0mm 3mm 0mm, clip]{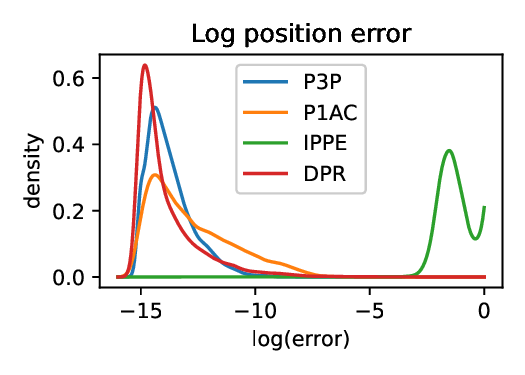}
    \vspace{.05cm}
    \caption{\label{subfig:stability}Numerical stability}

    \end{subfigure}
    \begin{subfigure}[b]{0.7\textwidth}
    \centering
      \includegraphics[height=3.75cm]{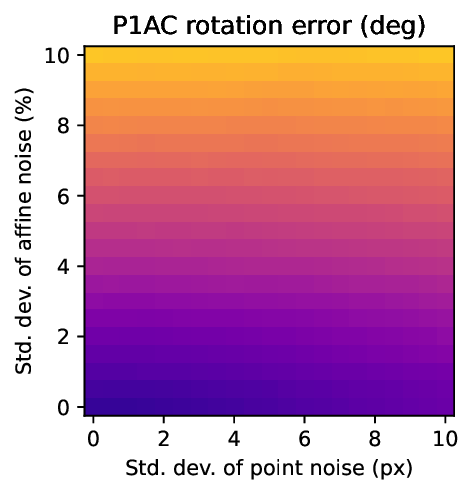}
      \includegraphics[height=3.75cm]{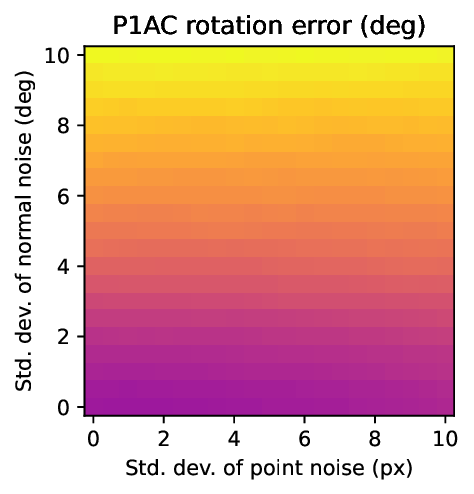}
      \includegraphics[height=3.75cm]{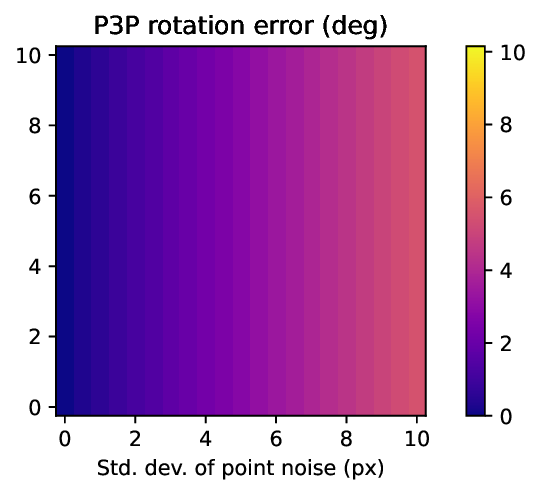}\\
      \includegraphics[height=3.75cm]{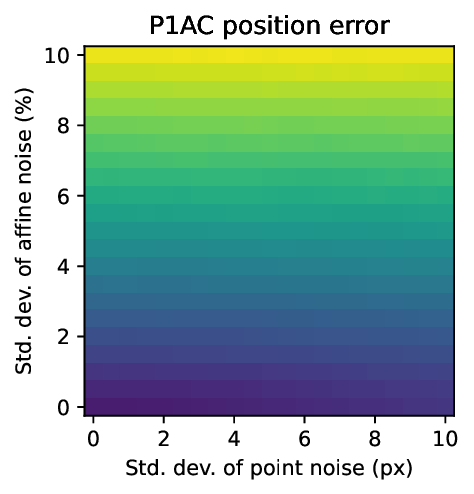}
      \includegraphics[height=3.75cm]{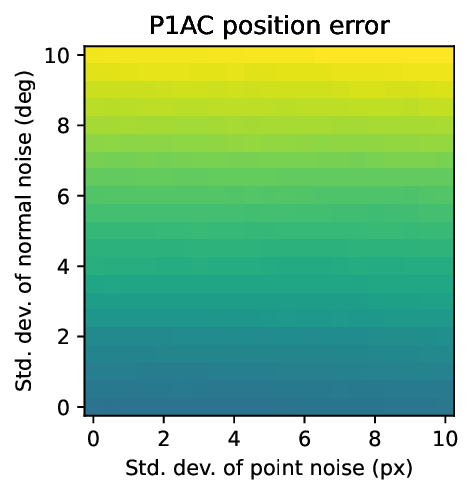}
      \includegraphics[height=3.75cm]{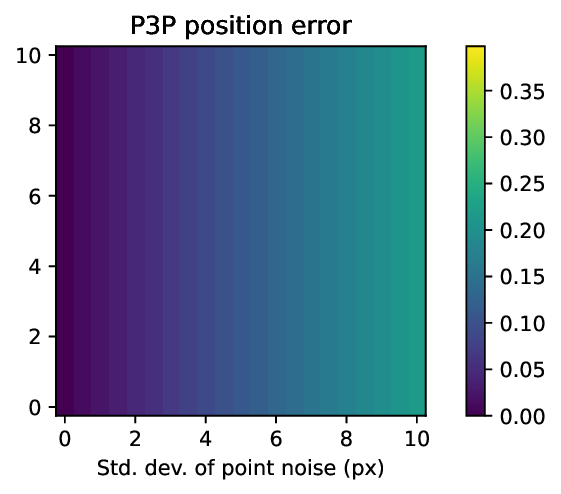}    
      \caption{\label{subfig:noise}Noise}
      
      \end{subfigure}
    \caption{\label{fig:stability_and_noise}Results of synthetic data experiments. 
 (a) Analysis of numerical stability with zero noise added to observations. The plots are estimates of the distribution produced by Gaussian kernel-density estimation.  \emph{Top:} Log rotation error; \emph{Bottom:} Log position error. 
 (b) Analysis of error with respect to various types and levels of noise. \emph{Left:} Error of P1AC solver w.r.t. affine and point noise, with normal noise fixed to 1$^\circ$.  \emph{Center:} Error of P1AC solver w.r.t. normal and point noise, with affine noise fixed to 4\%. \emph{Right:} Error of P3P solver w.r.t. affine and point noise. Note that the P3P solver does not use the affine transformation or normal vector and thus is unaffected by affine and normal noise.
 }
\end{figure*}
\begin{figure}
    \centering
    \includegraphics[width=\columnwidth]{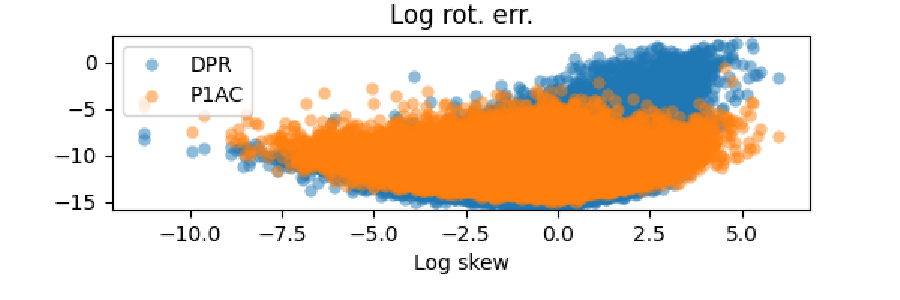}
    \caption{\label{fig:skew}Rotation accuracy versus skew in the affine transformation, over 1M random problem instances. The accuracy of the DPR solver degrades as perspective distortion increases, whereas our P1AC solver is stable.}
\end{figure}

\label{sec:evaluation}

We performed evaluations on synthetic and real data to test the numerical stability of our solver, test robustness to various types of noise, and assess its performance as part of an image-based localization system.

\noindent
We compared the following algorithms in our evaluation:
\begin{itemize}
\vspace{-1.0ex}
    \item \textbf{P3P} Standard P3P absolute pose calculation from three 2D-3D point observations.  We used the Lambda Twist P3P solver \cite{persson2018lambda}.
    \vspace{-1.6ex}
    \item \textbf{P1AC}: Our solution to the P1AC problem.
    \vspace{-1.6ex}
    \item \textbf{DPR}: Differential Pose Resection \cite{koser2008differential}.
    \vspace{-1.6ex}
    \item \textbf{IPPE}: Infinitesimal Plane Pose Estimation \cite{collins2014infinitesimal}.
\end{itemize}

In the experiments, we computed two pose accuracy metrics. Given the true camera rotation $\Matrix{R}$ and the estimated camera rotation $\Matrix{\hat{R}}$ we computed the \emph{rotation error} as $||\ln(\Matrix{\hat{R}}\Matrix{R}^\text{T})||$, where $\ln(\cdot)$ denotes the SO(3) logarithm \cite{huynh2009metrics,hartley2013rotation}.  While $\textrm{arccos}((\textrm{trace}(\Matrix{R})-1)/2)$ is often used as a rotation error metric, we instead use a robust implementation of $||\ln(\Matrix{\hat{R}}\Matrix{R}^\text{T})||$, which we found to be more stable for small rotations.  Given the true camera center $\Vector{c}$ and the estimated camera center $\Vector{\hat{c}}$, we computed the \emph{position error} as $||\Vector{c}-\Vector{\hat{c}}||$.  In the synthetic experiments, when considering multiple solutions produced by a method, we choose the solution that minimizes the max.\ of these two error metrics.

\subsection{Synthetic data experiments}
We generated synthetic problem instances with varying levels of noise added to the 2D point observations, affine transformations, and normal vectors.

We generated synthetic random problems in the following manner adapted from \cite{eichhardt2020relative}.  We randomly place two cameras around the origin at a distance sampled from $[1,2]$, both oriented towards a random target point sampled from $[-0.5,0.5]^3$.  To generate a correspondence between the two cameras, we select a random 3D point from $\mathcal{N}(\Vector{0},\Matrix{I}_{3\times3})$ with a random normal vector.  We project the 3D point to the two cameras using a focal length of $400$.  We then calculate the affine transformation from the local homography between the first and second cameras \cite{barath2016novel}.  We transform the cameras and 3D points so that the first camera (the reference camera) is at the origin with identity rotation.  

We add Gaussian noise to the 2D point observations.  Following prior studies \cite{eichhardt2018affine,eichhardt2020relative}, we add Gaussian noise to the elements of the $2\times2$ affine transformation.  
We select the level of affine noise based on the chosen percentage error; i.e., if the chosen percentage error is $e$, the std.~dev.~of the noise added to $a_{ij}$ is $e \cdot a_{ij}$. 
Finally, we add noise to the normal vectors by rotating them by a random rotation with angles randomly chosen from a Gaussian distribution.  For each problem instance, we generate three correspondences between the two cameras, so that we can test the 1AC solvers (using the first correspondence) and P3P (using all three correspondences) on the same problem instance.

\PAR{Numerical stability.}
To evaluate the numerical stability of each solver, we computed the rotation and position errors of each solver over 10,000 random problem instances with no added observational noise.  Figure \ref{subfig:stability} shows the results.  
In the context of the minimal solver literature, the primary objective of such stability tests is \textit{not} to facilitate comparisons between different solvers but to verify the individual stability of each solver, which is indicated by an absence of peaks above $10^{-2}$ to $10^{-1}$.
It is important to note that $> 99.9\%$ of the errors for P3P and P1AC are located below $10^{-5}$. 
Practically speaking, this suggests these solvers would yield usable results. 
IPPE has substantially worse performance, as expected due to the unavoidable error in the virtual correspondences used to calculate IPPE. 

To further investigate the difference between P1AC and DPR, we analyzed the performance of each solver with respect to the amount of perspective distortion between the reference and query images.  To quantify the distortion, we applied a QR decomposition of the affine matrix $\Matrix{Q}\Matrix{R} = \Matrix{A}$ such that $\Matrix{Q}$ is a 2D rotation matrix and $\Matrix{R}$ is upper-triangular.  The value at $R_{12}$ represents the amount of skew in the transformation, and indicates the level of perspective distortion.  
Figure \ref{fig:skew} shows a scatter plot of rotation accuracy versus skew for both solvers over 1M random problem instances.  DPR has a noticeable increase in error as skew increases, whereas P1AC is unaffected.  This increase in error is caused by the approximate warping of the AC required for the DPR method.

\PAR{Timings.}
We computed the average processing times of the solvers over 10,000 random problem instances.  
The solvers were all implemented in C++ and timings were made on an Apple M1 Pro CPU.  We used the implementations of Lambda Twist and 3Q3 provided in PoseLib \cite{PoseLib} and the implementation of IPPE provided in OpenCV's solvePnP function.  The results are given in Table \ref{table:timings}.  

%\vspace{-1.0ex}
\begin{table}
\begin{center}
\begin{tabular}{|l| c | c c c |}
\hline
 % &  & \multicolumn{3}{c|}{P1AC} \\
 %& P1AC (Null) & (Elim) &  (3Q3) \\
& P3P & P1AC & DPR & IPPE \\
\hline
% Time ($\mu$s)  & 1.18 & 53.94 & 4.74 \\
Time ($\mu$s)  & 0.54 & 2.73 & 3.23 & 25.27 \\ % new timings from Apple M1 Pro and eliminated solver
\hline
\end{tabular}
\end{center}
\vspace{-1.8ex}
\caption{\label{table:timings}Average timing in $\mu$s over 10,000 trials. }%Although the nullspace solver is slower than P3P, it reduces overall image localization time by reducing the number of RANSAC iterations.}
\end{table}

Among the 1AC variants, P1AC is the fastest with an average timing of 2.73 $\mu$s.  The Lambda Twist P3P solver is about $6\times$ faster than P1AC.   Note that this implementation of the P3P solver is based on a careful elimination of the rotation and translation parameters to reveal the underlying elliptic equations in the P3P problem which are solved through diagonalization.

\PAR{Noise.}
We tested the P1AC solver and the P3P solver over a range of noise levels in the 2D point observations, affine transformations, and surface normals.  We tested 1,000 random problem instances for each noise level.  Figure \ref{subfig:noise} plots the median errors for each method at each setting.  The chosen noise ranges are the same as used in \cite{eichhardt2020relative}.
The left column  presents the results for increasing levels of point and affine noise, with the normal noise set to 1 degree. The center column presents the results for increasing levels of point and normal vector noise, with the affine noise set to 4\% \cite{eichhardt2020relative}.  The right column presents the results for the P3P solver, for comparison.  The P3P solver does not use the affine transformation or normal vector and thus is unaffected by affine and normal noise.  

The synthetic experiments show
that the P1AC solver is affected by affine and normal noise much more than by 2D point observation noise.
In the presence of affine and normal noise, it returns larger errors than the P3P solver, which is not affected by these noises. Nevertheless, for noise levels that correspond to realistic values, i.e., 2-4\% for affine
noise and $1^\circ$ for normal noise, the P1AC solver still returns reasonably small errors that are sufficient for the initialization of local optimization~\cite{Lebeda2012BMVC} inside RANSAC.

\PAR{Robust estimation.}
We investigated the performance of P1AC and P3P when used in robust estimation with an increasing outlier ratio, and confirmed that within a robust estimation loop, P1AC outperforms P3P in terms of speed.   

We tested a range of outlier ratios from 0 to 0.9, with 10,000 trials at each setting.  In each trial we randomly generate 1,000 correspondences with a point noise set at 1 pixel, affine noise at 4\%, and normal noise at $1^\circ$. We replace a portion of the correspondences with random values according to the desired outlier ratio.  We then run 
vanilla RANSAC \cite{fischler1981random} or LO-RANSAC \cite{Lebeda2012BMVC} using either P1AC or P3P as the minimal solver and EPnP \cite{lepetit2009epnp} as the non-minimal solver in local optimization.  We tested vanilla RANSAC to analyze the trade-off in our approach between having a smaller minimal sample size but greater noise sensitivity.  LO-RANSAC mitigates noise sensitivity in the minimal solver by incorporating local optimization (LO) iterations to grow the inlier set from the initial estimate.

The average error was stable for each method across the range of outlier ratios.  Table \ref{tab:ransac_error} shows the average error for each method across all outlier ratios.  When employing vanilla RANSAC, P1AC is less accurate than P3P at this level of noise.
When employing LO-RANSAC, while the error for P1AC is marginally higher than that for P3P, the error rate for both methods remains extremely low, attesting to their effectiveness.  Note that on real data, P1AC substantially outperforms P3P (see Section \ref{sec:realdata}). 

\begin{table}
    \centering
    \resizebox{0.9\linewidth}{!}{
    \begin{tabular}{|l|r r|r r|}
        \hline
         ~ & \multicolumn{2}{c|}{P1AC} & \multicolumn{2}{c|}{P3P} \\
         Robust estimator & Rot. ($^\circ$) & Pos. & Rot. ($^\circ$) & Pos. \\
         \hline
         \hline
         RANSAC    & 0.450 & 0.019 & 0.056 & 0.002 \\
         LO-RANSAC & 0.085 & 0.003 & 0.019 & 0.001 \\
         \hline
    \end{tabular}
    }
    \caption{Average rotation and pose error for each robust estimation method on synthetic data over a range of outlier ratios.}
    \label{tab:ransac_error}
\end{table}

\begin{figure}
    \centering
    \includegraphics[width=0.5\textwidth]{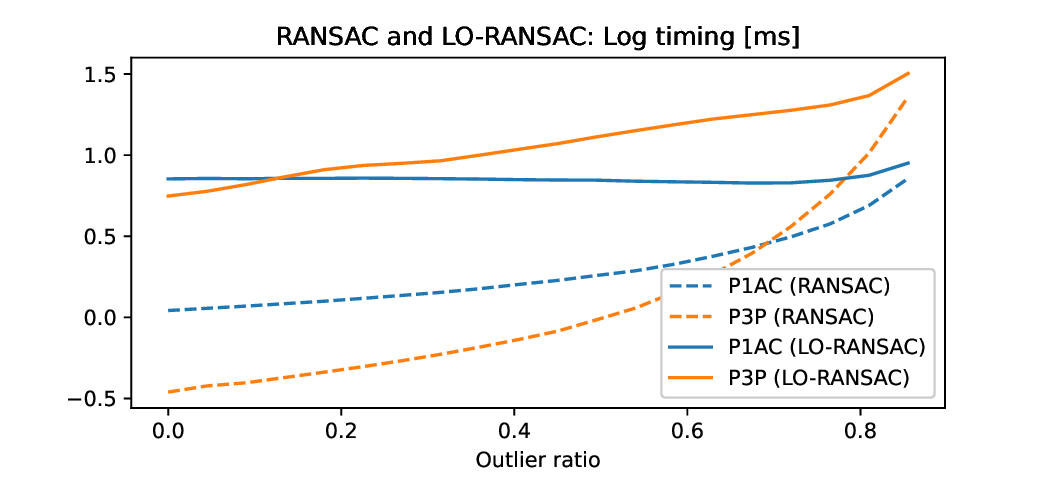}
    \caption{Average timing for each robust estimation method across a range of outlier ratios.}
    \label{fig:ransac_timings}
\end{figure}

Figure \ref{fig:ransac_timings} plots the average timing for each method.   When using vanilla RANSAC, P1AC is faster than P3P past an outlier ratio of about 0.7; with LO-RANSAC, P1AC is faster past an outlier ratio of about 0.1. This result indicates that our P1AC solver is accurate enough to provide a good initialization for LO and reach convergence more quickly than P3P, even with a low outlier ratio. 
In our real data experiments (Section \ref{sec:realdata}), to be consistent with the state-of-the-art, we use Graph Cut RANSAC (GC-RANSAC) \cite{barath2021graph}, which improves upon LO-RANSAC by incorporating graph cut optimization into the LO iterations.  We did not use GC-RANSAC in our synthetic data experiments because the observations, being randomly generated, did not exhibit spatial coherence.

\PAR{Mirrored solutions and degeneracies.}
The solutions returned by our P1AC solver include mirrored solutions, meaning pairs of solutions where the cameras are rotated exactly 180 degrees from each other.  
Previous work on plane-based pose estimation has also mentioned the existence of mirrored solutions \cite{collins2014infinitesimal}; they are indeed common for minimal problems and are an artifact of not including inequalities in the constraints.  In practice they are not a problem since such solutions are filtered out inside RANSAC.

In addition to the 180$^\circ$ degeneracy inherent to the Cayley rotation representation, we noticed a degeneracy for rotations very close to identity when using the PoseLib 3Q3 solver implementation \cite{PoseLib}.  However, when using a 3Q3 solver produced by the GAPS automatic generator \cite{li2020gaps}, this degeneracy for near-identity rotations does not appear.  Another observation we made is that the translation accuracy of both P1AC and P3P degrades in weak-perspective situations, where the plane is very far from the camera.  These aspects are discussed further in the supplemental material.

\subsection{Real data experiments}
\label{sec:realdata}

\begin{table*}
\begin{center}
\resizebox{1.0\textwidth}{!}{
\begin{tabular}{| l | r r r r | r r r r | r r r r | r r r r | r r r r | }
\hline
& \multicolumn{4}{|c|}{Position (cm) $\downarrow$ }
& \multicolumn{4}{|c|}{Rotation ($^\circ$) $\downarrow$ }
& \multicolumn{4}{|c|}{Recall (0.1m/$1^\circ$) $\uparrow$}
& \multicolumn{4}{|c|}{Recall (0.2m/$1^\circ$) $\uparrow$} \\
Scene            & P3P & DPR & IPPE & P1AC & P3P & DPR & IPPE & P1AC & P3P & DPR & IPPE & P1AC & P3P & DPR & IPPE & P1AC \\
\hline\hline
     Great Court &
     8 &        \textbf{6} &          7 &          \textbf{6} &
     0.1 &     \textbf{0.08} &        0.1 &       \textbf{0.08} &
     61.8 &     \textbf{74} &       65.3 &        73.6 &
     86.4 &          90.1 &       86.2 &        \textbf{91.1} \\
  King's College &
  10 &        8 &          9 &          \textbf{7} &
  0.16 &      \textbf{0.14} &       0.2 &       \textbf{0.14} &
  49.3 &          \textbf{65.3} &       57.7 &        64.7 &
  79.3 &          84.9 &       85.4 &        \textbf{86.9} \\
    Old Hospital &
    9 &        \textbf{7} &          8 &          \textbf{7} & 
    0.15 &      0.12 &       0.14 &       \textbf{0.11} &
    58.8 &          68.9 &       62.6 &        \textbf{72.5} &
    79.7 &          89.6 &       85.2 &        \textbf{93.4} \\
     Shop Façade &
     4 &        \textbf{3} &          4 &          \textbf{3} &
     \textbf{0.13} &      0.15 &       0.14 &       \textbf{0.13} &
     85.4 &          86.4 &       84.5 &        \textbf{87.4} &
     94.2 &          93.2 &       92.2 &        \textbf{95.1} \\
St Mary's Church &
7 &        \textbf{5} &          6 &          \textbf{5} &
0.21 &      \textbf{0.17} &        0.2 &       \textbf{0.17} &
61.7 &          70.1 &       67.6 &        \textbf{72.5} &
80.9 &          85.2 &       84.3 &        \textbf{87.7} \\
          Street &
          32 &       50 &        248 &         \textbf{20} &
          0.57 &       0.8 &       \textbf{0.25} &       0.35 &
          10.7 &          16.2 &       19.1 &        \textbf{19.8} &
          18 &            24 &       26.7 &        \textbf{29.7} \\
\hline
   Avg. &
   12 &       13 &         47 &          \textbf{8} &
   0.22 &      0.24 &       \textbf{0.16} &       \textbf{0.16} &
   54.62 &         63.48 &       59.5 &       \textbf{65.08} &
   73.08 &         77.83 &       76.7 &       \textbf{80.65} \\
  Weighted avg. &
  22 &       33 &        153 &         \textbf{14} &
  0.40 &      0.53 &       \textbf{0.21} &       0.26 &
  30.44 &         38.13 &       37.4 &       \textbf{40.62} &
  43.91 &         49.33 &       50.1 &       \textbf{53.53} \\
\hline
\end{tabular}
% \vspace{-12pt}
}
\end{center}
\caption{\label{table:cambridgeerror}
\textbf{Cambridge Landmarks}~\cite{kendall2015posenet}
median position (centimeters) and rotation (degrees) errors, and recalls (percentages), at 0.1m/1$^\circ$ and 0.2m/1$^\circ$, of GC-RANSAC~\cite{barath2021graph} combined with various solvers.
The average over all scenes and average weighted by number of images in each scene are in the last row.
}
\end{table*}

\begin{table}
\begin{center}
\resizebox{0.7\linewidth}{!}{\begin{tabular}{| l | c c c c |  }
\hline
 & \multicolumn{4}{|c|}{Recall (0.25m/2$^\circ$) $\uparrow$}\\
 & P3P & DPR & IPPE & P1AC \\
\hline\hline
% Day & 61.3 & \textbf{61.9} & 83.4 & \textbf{84.0} & 95.1 & \textbf{95.6} \\
%Night & 16.3 & \textbf{17.3} & 32.7 & \textbf{34.7} & 74.4 & \textbf{83.7} \\
Day & 62.0 & \textbf{62.4} & 62.0 & 62.0 \\
Night & 47.1 & 47.6 & 46.1 & \textbf{51.3}  \\
\hline
 & \multicolumn{4}{|c|}{Recall (0.5m/5$^\circ$) $\uparrow$}  \\
\hline
Day & 83.4 & 84.3 & 83.4 & \textbf{84.6} \\
Night & 60.2 & 64.9 & 63.9 & \textbf{66.0} \\
\hline
 & \multicolumn{4}{|c|}{Recall (5m/10$^\circ$) $\uparrow$} \\
 \hline
Day & \textbf{96.0} & 95.6 & \textbf{96.0} & 95.9 \\
Night & 74.3 & 75.9 & 76.4 & \textbf{82.2} \\
\hline
\end{tabular}}
\end{center}
\caption{\label{table:aachenerror} \textbf{Aachen Day-Night} pose error recalls~\cite{sattler2012image}, in percentages, at 0.25m/2$^\circ$, 0.5m/5$^\circ$, and 5.0m/10$^\circ$ of GC-RANSAC~\cite{barath2021graph} combined with various solvers.}%P3P and the proposed P1AC (3Q3) minimal solver. }
\end{table}

To evaluate the performance of our method for large-scale image-based localization, we used the  Cambridge Landmarks~
\cite{kendall2015posenet} and Aachen Day-Night v1.1~\cite{sattler2012image,Sattler2018CVPR,Zhang2020IJCV} benchmark datasets. 
Both datasets are commonly used in the image-based localization literature. 

The Cambridge Landmarks dataset consists of six scenes, each recorded through multiple video sequences taken by a mobile phone, depicting parts of Cambridge, UK. 
Out of the sequences recorded for each dataset, some are used to obtain database images that represent the scene while the others are used to obtain query images. 
Ground truth poses and intrinsic camera calibrations for all images were obtained using the VisualSFM~\cite{wu2013towards,wu2011multicore} Structure-from-Motion (SfM) software. 
Image-based localization performance is typically measured by reporting the median position and orientation error. 
In addition, we measure the percentage of images localized within 10cm/1$^\circ$ and 20cm/1$^\circ$ of their ground truth poses. 

Cambridge Landmarks is a relatively easy dataset since each scene is rather small and the query images were taken around the same time as the database images. 
A much more challenging case is the Aachen Day-Night dataset, which depicts the historical inner city of Aachen, Germany. 
The database images were all taken during daytime conditions and the dataset provides query images taken at nighttime conditions over a longer period of time. 
Ground truth for the database images were obtained via COLMAP~\cite{schoenberger2016sfm}. 
The nighttime queries were later registered by refining initial pose estimates~\cite{Zhang2020IJCV}. 
We follow the common evaluation protocol and report the percentage of images localized within three error thresholds (0.25m/2$^\circ$, 0.5m/5$^\circ$, 5.0m/10$^\circ$).

There are multiple ways to obtain affine features from real images. 
The most standard is to use a local feature detector, like DoG~\cite{lowe2004distinctive} or Key.Net~\cite{KeyNet2019}, estimate keypoint locations and scales, and use the patch-based AffNet~\cite{mishkin2018repeatability} to get affine shapes. 
Finally, a patch-based descriptor, like HardNet~\cite{HardNet2017} or SOSNet~\cite{SoSNet2019}, is applied. 
In our experiments, we combine DoG for feature detection, AffNet for affine shape recovery, and HardNet for patch description, an approach that is
among the leaders in the IMC 2020 benchmark~\cite{IMC2020}.
Given a SfM point cloud, we apply the normal estimator implemented in MeshLab to obtain an oriented point cloud. 
The normals are fitted to the $200$ nearest neighbors. 

\PAR{Image localization procedure.}
For both datasets, we use image retrieval to identify a small set of potentially relevant database images for each query photo. 
We use DenseVLAD~\cite{Torii2015CVPR} descriptors for Cambridge Landmarks and NetVLAD~\cite{Arandjelovic2016CVPR} descriptors for Aachen Day-Night. 
We then establish 2D-2D correspondences between each query and its retrieved database images. 
These 2D-2D matches are then lifted to 2D-3D correspondences. 
For Cambridge Landmarks, we use the 3D points in SfM point clouds corresponding to the features extracted from the database images. 
For Aachen, we use a dense depth maps obtained by rendering a mesh model of the scene to obtain the 3D points (and their corrresponding normals) for the database features~\cite{Panek2022ECCV}.

In these experiments we aim to analyze the usefulness of the P1AC solver in practice.  To this end we use advanced RANSAC techniques and tune them per solver, to achieve the best performance possible for each method.

To estimate the pose of each query image, we iterate through the retrieved database images.
Between each pair, we estimate the pose by GC-RANSAC~\cite{barath2021graph} using the implementation provided by the authors.
The final pose for a query image comes from the database image with which the highest inlier number is achieved. 
GC-RANSAC requires two types of solvers, one for minimal and one for non-minimal pose estimation. 
We test the proposed P1AC solver as the minimal solver and compare it to the P3P solver~\cite{persson2018lambda}.
The non-minimal solver, running both in the local optimization and in the final refitting on all inliers, is always EPnP~\cite{lepetit2009epnp} followed by a few iterations of the Levenberg-Marquardt optimization~\cite{more1978levenberg} minimizing the reprojection error.
The proposed P1AC solver enables running GC-RANSAC exhaustively, without severely affecting the run-time.
Therefore, we perform an exhaustive search over all correspondences without using a random sampler. 
For P3P solver, we use the PROSAC sampler~\cite{chum2005matching} on matches ordered by SNN ratio. 

For Cambridge Landmarks, we set the inlier-outlier threshold to $4$ pixels.
For Aachen Day-Night, we set it to $12$ pixels.
We tuned the SNN ratio threshold for both P3P and P1AC. For P3P, $0.9$ works the best. 
Surprisingly, P1AC and other 1AC solvers are the most accurate on Cambridge Landmarks without any SNN ratio filtering. 
Even though this leads to slightly increased run-times due to running on much more points, it increases both the accuracy and inlier number significantly.

\PAR{Pose accuracy.}
A summary of errors on the Cambridge Landmarks dataset is given in Table \ref{table:cambridgeerror}.  We report the median position (in meters) and rotation  errors (in degrees), and the recall (in percentages) at 0.1m/1$^\circ$ and 0.2m/1$^\circ$.
The proposed P1AC method has lower errors and thus, higher recall on \textit{all} scenes.
In many cases, the recall increase is significant compared to the P3P solver.
For example, on scene Street (the most challenging scene in the dataset) both recall values increase by around 10 percentage points (pp). 
The average 0.1m/1$^\circ$ over all scenes is increased by 10.5pp and the average 0.2m/1$^\circ$ by 7.6pp.
Compared to DPR and IPPE, the proposed P1AC solver increases the recall by 1-3pp on all scenes, while significantly reducing the median position error. 

The recalls at 0.25m/2$^\circ$, 0.5m/5$^\circ$, and 5.0m/10$^\circ$ on the Aachen Day-Night~\cite{sattler2012image} dataset are in Table~\ref{table:aachenerror}.
In this case, all methods use an SNN ratio threshold set to $0.9$ since increasing it led to a deterioration in the accuracy.
This is likely caused by the dataset being more challenging than the Cambridge Landmarks.
While the accuracy difference is smaller than in Table \ref{table:cambridgeerror}, the proposed P1AC consistently improves all metrics on the Night sequence and leads to similar accuracy to all other methods in the Day sequence.

\PAR{Timing.}
Table~\ref{table:cambridgetimings} reports the average processing times (in seconds) of all compared methods.
On Cambridge Landmarks, as expected from the excluded SNN ratio filtering and the exhaustive search used for 1AC methods, their run-time is larger than that of P3P.
On Aachen Day-Night, all methods use the same SNN ratio threshold and the proposed P1AC is faster on both scenes than P3P. 
P1AC is the fastest on average on the Day sequence, and runs at a comparable speed to other methods in the Night sequence.

\begin{table}
\begin{center}
\begin{tabular}{|l|r r r r|}
\hline
& \multicolumn{4}{|c|}{Time (secs) $\downarrow$} \\
Scene & P3P & DPR & IPPE & P1AC \\
\hline\hline
Great Court & \textbf{0.14} & 0.17 & 0.21 & 0.26 \\
King's College & \textbf{0.19} & 0.58 & 0.37 & 0.48 \\
Old Hospital & \textbf{0.12} & 0.59 & 0.39 & 0.42 \\
Shop Facade & \textbf{0.13} & 0.52 & 0.36 & 0.37 \\
St Mary's Church & \textbf{0.12} & 0.45 & 0.30 & 0.38  \\
Street & \textbf{0.10} & 0.37 & 0.28 & 0.24 \\
\hline
Day & 0.24 & 0.18 & 0.17 & \textbf{0.09}  \\
Night & 0.12 & \textbf{0.05} & 0.07 & 0.09 \\
\hline
\end{tabular}
\end{center}
\caption{
Avg.~%\ inlier numbers and
times of GC-RANSAC~\cite{barath2021graph} on the Cambridge Landmarks \cite{kendall2015posenet} and Aachen Day-Night~\cite{sattler2012image} datasets.}
\label{table:cambridgetimings}
\end{table}

%& \multicolumn{2}{|c|}{Inliers $\uparrow$} & \multicolumn{2}{|c|}{Time (secs) $\downarrow$} \\
%Scene & P3P & P1AC & P3P & P1AC \\
%\hline\hline
%Great Court & 533 & \textbf{772} & \textbf{0.14} & 0.26 \\
%King's College & 2679 & \textbf{3306} & \textbf{0.19} & 0.48 \\
%Old Hospital & 1064 & \textbf{1581} & \textbf{0.12} & 0.42 \\
%Shop Facade & 1578 & \textbf{2157} & \textbf{0.13} & 0.37 \\
%St Mary's Church & 1211 & \textbf{1536} & \textbf{0.12} & 0.38  \\
%Street & 209 & \textbf{390} & \textbf{0.10} & 0.24 \\
%\hline
%Day & 244 & \textbf{246} & 0.24 & \textbf{0.09}  \\
%Night & \textbf{56} & 52 & 0.12 & \textbf{0.09} \\

In our experiments, we used the same set of matches (DoG+HardNet+AffNet) for all methods, with the P3P method disregarding the affine components. As such, no method had a timing advantage related to feature extraction.  However, for completeness, we measured feature extraction time and found that DoG+HardNet needs $0.18$ ms per feature, and DoG+HardNet+AffNet $0.53$ ms.   Note that AffNet plays a crucial role in maintaining high accuracy and is not solely used for generating affine features \cite{IMC2020}.

\subsection{Discussion}
\label{sec:discussion}

In terms of the timing of the minimal solver by itself, our P1AC solver is slower than the P3P solver.  
However, our P1AC solver is still quite fast ($<3 \mu$s) and moreover the minimal solver computation time is usually negligible when compared to other steps in the random sampling process such as local optimization and inlier testing.

While our P1AC solver itself is slightly slower than P3P, and is susceptible to more types of noise, it has a number of beneficial properties due to estimating the pose from a single AC, such as deterministic and fast robust estimation. On real data, we show that, when employed inside a robust estimator, our P1AC solver produces more accurate results and also is faster in some cases (e.g., Aachen Day-Night).

In comparison to alternative baseline methods, DPR and IPPE, our P1AC method successfully localizes more frames and is more accurate.  This indicates that our proposed approach, based on differential analysis of the AC, more accurately models the constraint imposed by an AC on the absolute pose of the query camera.

Improving the AC quality \cite{barath2020making} might lead to a speed up in sample consensus because the initial pose estimates by the P1AC solver would be more accurate, so the system would spend less time on LO steps.  Other minimal solvers that use a single correspondence often use 
histogram or kernel density voting in place of RANSAC, although they typically estimate a single parameter such as the focal length \cite{bujnak2009robust,hajder2020relative}.  Future work lies in analyzing how ACs and the P1AC constraints could be applied to 6DOF camera pose voting \cite{zeisl2015camera,aiger2021efficient}.
While we only use point reprojection error for inlier testing due to the inherent noise in the affine components \cite{barath2020making}, future work could include testing first-order compatibility.

\section{Conclusions}
\label{sec:conclusion}
In this work, we derived novel constraints imposed by an affine correspondence on the absolute pose of a camera, given knowledge of a 3D point in the scene, the surface normal at that point, and the pose of the reference image.  Using these novel constraints we developed a minimal solution for absolute pose from a single AC.  Through experiments on synthetic data we established the numerical stability of our solver and its behavior under increasing levels of noise.We demonstrated the usefulness of our solver in an image-based localization system.  In our experiments, replacing P3P with our P1AC solver improved localization accuracy and recall.

{\small
\PAR{Acknowledgements.}
This work was supported by NSF Award No.~2144822, EU Horizon 2020 project RICAIP (No. 857306), the European Regional Development Fund under project IMPACT (No. CZ.02.1.01/0.0/0.0/15\_003/0000468),and the Czech Science Foundation (GAČR) JUNIOR STAR Grant No. 22-23183M.  D.~Barath was supported by the ETH Postdoc Fellowship.
}

{\small
\bibliographystyle{ieee_fullname}
\bibliography{references}
}

\end{document}

\newcommand{\Vector}[1]{\lowercase{\mathbf{#1}}}
\newcommand{\Matrix}[1]{\uppercase{\mathsf{#1}}}
\newcommand{\Skew}[1]{[\Vector{#1}]_\times}

\newcommand{\TS}[1]{\textcolor{blue}{{TS: [#1]}}}
\newcommand{\JV}[1]{\textcolor{red}{{JV: [#1]}}}

%%%%%%%%% TITLE
\title{--- Supplementary Material ---\\ P1AC: Revisiting Absolute Pose From a Single Affine Correspondence}

\author{Jonathan Ventura$^1$ \quad Zuzana Kukelova$^2$ \quad Torsten Sattler$^3$ \quad D\'{a}niel Bar\'{a}th$^4$\\
$^1$ Department of Computer Science \& Software Engineering, Cal Poly\\ 
$^2$ Visual Recognition Group, Faculty of Electrical Engineering, Czech Technical University in Prague\\  
$^3$ Czech Institute of Informatics, Robotics and Cybernetics, Czech Technical University in Prague\\
$^4$ Department of Computer Science, ETH Z\"{u}rich
}

\maketitle
\ificcvfinal\thispagestyle{empty}\fi

%%%%%%%%% ABSTRACT
%%%%%%%%% BODY TEXT
In this supplementary material, we present additional experiments to support the claims in the main paper, explain and analyze alternative solver formulations, and present more detailed explanations of some aspects of the main paper. Section \ref{sec:noise} presents extra experimental results on noise sensitivity. Section \ref{sec:ippe} evaluates an alternative application of the IPPE method.
Section \ref{sec:degeneracies} analyzes degenerate cases for the P1AC solver.  Section \ref{sec:linearized} introduces a fast P1AC solution using an approximate rotation representation which is only valid for small rotations.  Section \ref{sec:nullspace} details an alternate P1AC solution using the $3\times3$ rotation matrix representation. 

\section{Noise sensitivity tests}
\label{sec:noise}

\begin{figure}[h!]
    \centering
    \begin{subfigure}[b]{8cm}
    \centering
      \includegraphics[height=5cm]{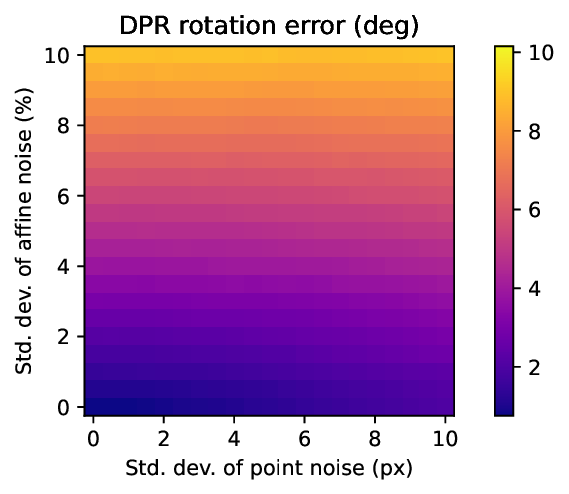}
      \includegraphics[height=5cm]{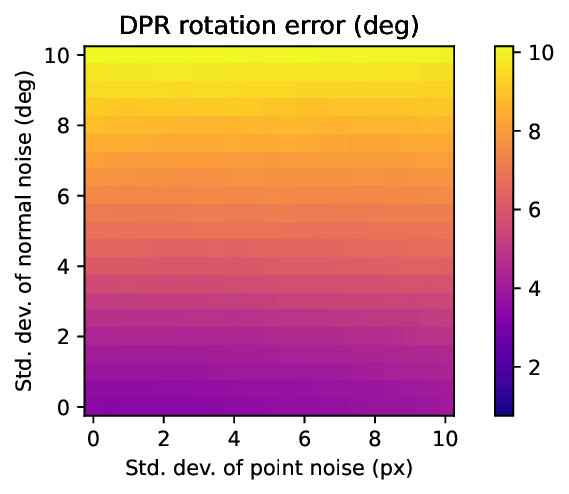}%\\
      % \includegraphics[height=3.75cm]{images/DPR_position_error_point_vs_affine_noise.eps}
      % \includegraphics[height=3.75cm]{images/DPR_position_error_point_vs_normal_noise.eps}
      % \caption{\label{subfig:noise}DPR Results w/ Noise}      
    \end{subfigure}
    \begin{subfigure}[b]{8cm}
    \centering
      \includegraphics[height=5cm]{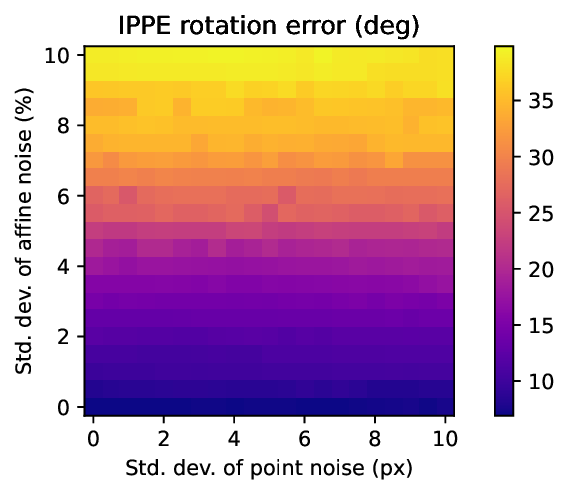}
      \includegraphics[height=5cm]{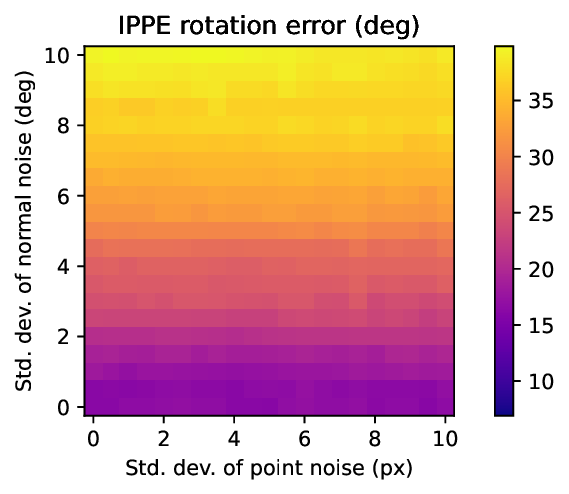}%\\
      % \includegraphics[height=3.75cm]{images/IPPE_position_error_point_vs_affine_noise.eps}
      % \includegraphics[height=3.75cm]{images/IPPE_position_error_point_vs_normal_noise.eps}
      % \caption{\label{subfig:noise}IPPE Results w/ Noise}      
    \end{subfigure}
\caption{\label{fig:noise_dpr_and_ippe}Analysis of error from DPR and IPPE methods with respect to various types and levels of noise. }
\end{figure}

Figure \ref{fig:noise_dpr_and_ippe} plots the rotation error versus noise for IPPE and DPR. The results for DPR are similar to P1AC, whereas IPPE's error is far higher (note the color bar range).

\section{IPPE with four point correspondences}
\label{sec:ippe}

\begin{table}
\begin{center}
% \resizebox{1.0\columnwidth}{!}{
% \begin{tabular}{| l | r r | r r | r r | r r | r r | }
\begin{tabular}{| l | c c | c c | }
\hline
% & \multicolumn{4}{|c|}{Position (cm) $\downarrow$ }
% & \multicolumn{4}{|c|}{Rotation ($^\circ$) $\downarrow$ }
& \multicolumn{2}{|c|}{Recall (0.1m/$1^\circ$) $\uparrow$}
& \multicolumn{2}{|c|}{Rec. (0.2m/$1^\circ$) $\uparrow$} \\
% Scene            & P1AC & IPPE (PC) & P1AC & IPPE (PC) & P3P & IPPE (PC) \\
Dataset            & P1AC & IPPE (4PC) & P1AC & IPPE (4PC) \\
\hline\hline
     Cambridge Landmarks & \textbf{65.08} & 46.51 & \textbf{80.65} & 58.92 \\
\hline
\end{tabular}
% }
\end{center}
\caption{\label{table:ippe_4pc_cambridge}
Comparison of P1AC and IPPE (4PC) results on \textbf{Cambridge Landmarks}.
}
% \vspace{-7mm}
\end{table}

\begin{table}
\begin{center}
% \resizebox{1.0\columnwidth}{!}{
% \begin{tabular}{| l | r r | r r | r r | r r | r r | }
\begin{tabular}{| l | c c | c c | c c | }
\hline
% & \multicolumn{4}{|c|}{Position (cm) $\downarrow$ }
% & \multicolumn{4}{|c|}{Rotation ($^\circ$) $\downarrow$ }
& \multicolumn{2}{|c|}{Recall (0.25m/$2^\circ$) $\uparrow$}
& \multicolumn{2}{|c|}{Recall (0.5m/$5^\circ$) $\uparrow$}
& \multicolumn{2}{|c|}{Recall (5m/$10^\circ$) $\uparrow$} \\
Dataset            & P1AC & IPPE (4PC) & P1AC & IPPE (4PC) & P1AC & IPPE (4PC) \\
\hline\hline
     % Aachen Day & \textbf{62.0} & 60.1 & \textbf{83.4} & 81.3 & \textbf{96.0} & 92.6 \\
     Aachen Day & \textbf{62.0} & 60.1 & \textbf{84.6} & 81.3 & \textbf{95.9} & 92.6 \\
     Aachen Night & \textbf{51.3} & 38.2 & \textbf{66.0} & 49.2 & \textbf{82.2} & 60.2 \\
\hline
\end{tabular}
% }
\end{center}
% \vspace{-10mm}
\caption{\label{table:ippe_4pc_aachen}
Comparison of P1AC and IPPE (4PC) results on \textbf{Aachen Day-Night}.}
\end{table}

We conducted extra tests of the IPPE solver on real data  using four PCs (rather than a single AC with added virtual correspondences) within the GC-RANSAC framework, in a manner consistent with our other experiments.  Given the requirement of IPPE for coplanar points, we implemented the NAPSAC sampler \cite{torr2002napsac} within GC-RANSAC to select proximate points in the samples.  As shown in Tables \ref{table:ippe_4pc_cambridge} and \ref{table:ippe_4pc_aachen}, the proposed P1AC substantially outperforms IPPE (4PC) on all tested datasets. 

\section{Degeneracies}
\label{sec:degeneracies}

As mentioned in Section 4.1 of the main paper, we investigated several specific problem configurations to search for situations where our P1AC solver might produce inaccurate results.

\subsection{Weak perspective}

\begin{figure}[h]
    \centering
    \includegraphics[width=0.45\textwidth]{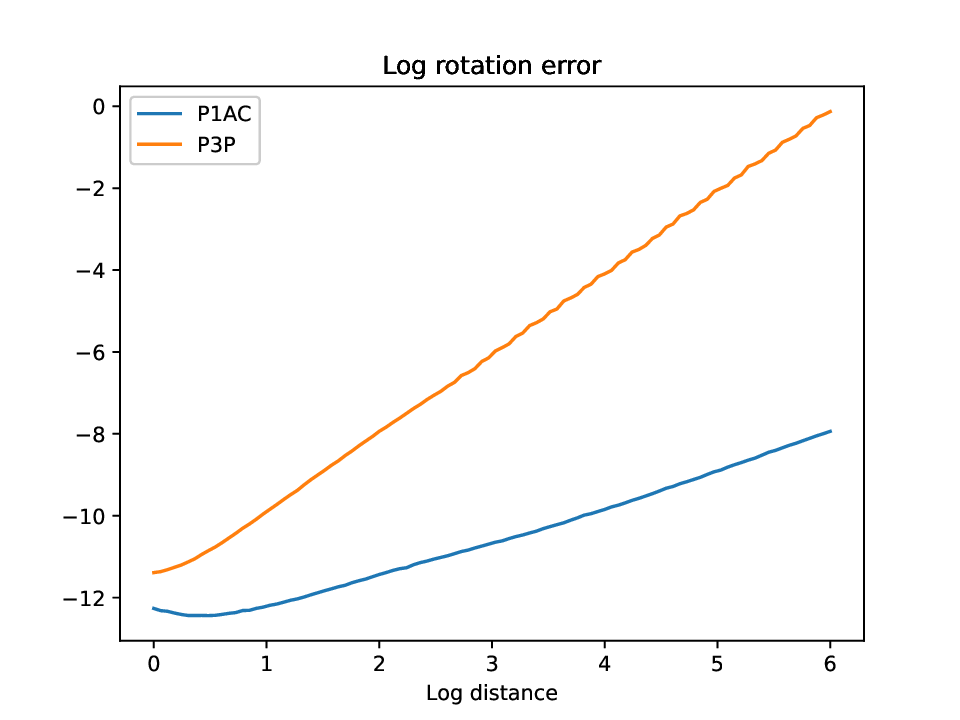}
    \includegraphics[width=0.45\textwidth]{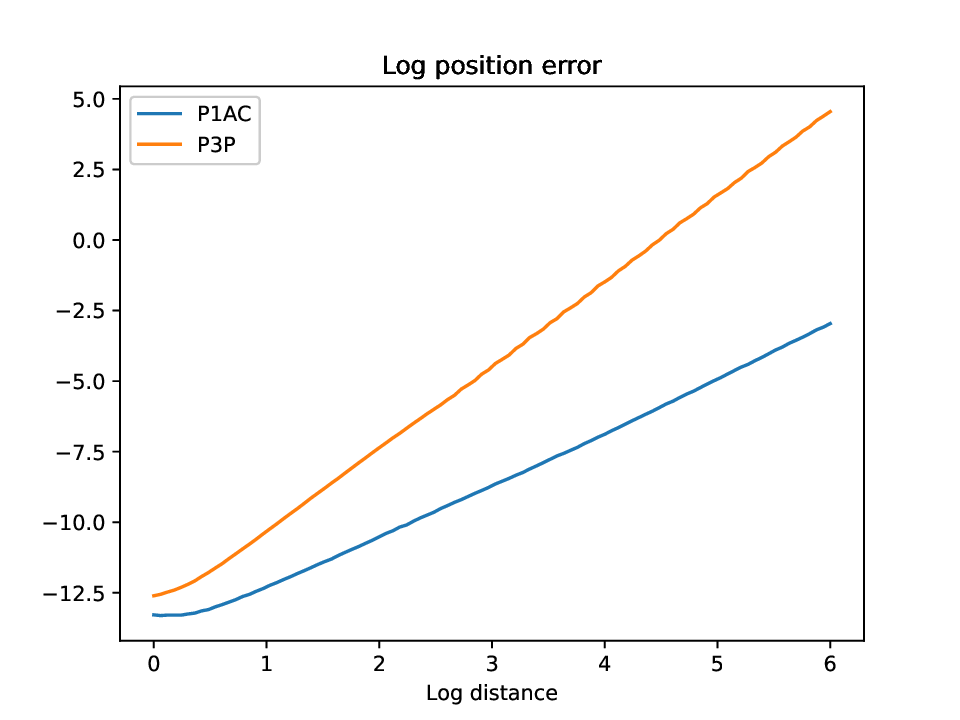}
    \caption{Plot of rotation and position error of P1AC and P3P over increasing distance of the query camera from the point. The error of both solvers increases with distance, although the error of P3P increases at a faster rate than P1AC.}
    \label{fig:distance}
\end{figure}

We investigated the change in performance as the configuration approaches weak perspective, where the differences in depths in the object / scene are much smaller than the distance to the object / scene.  To test this scenario, we generated random problems in the following manner.  We select a random rotation for the object plane and a center point for the plane selected from a 3D normal distribution.  Two co-planar points are placed at a distance of 0.1 from the center point along the principal axes of the plane.  A random reference camera is selected with unit distance from the origin, and a query camera at the chosen distance from the origin.  Both the reference and query camera look at a random target point sampled from $[-0.5~0.5]^3$.  No noise was added to the observations in this experiment.

As the distance of the query camera increases, the projected size of the planar object decreases, and the configuration approaches the weak perspective case.  Figure \ref{fig:distance} plots the rotation and position error of P1AC and P3P as the distance of the query camera increases.  The error of both P1AC and P3P increases with query camera distance, with the error of P3P increasing more quickly than P1AC.  

\subsection{Near-identity rotation}

\begin{figure}[h]
    \centering
    \includegraphics[width=0.45\textwidth]{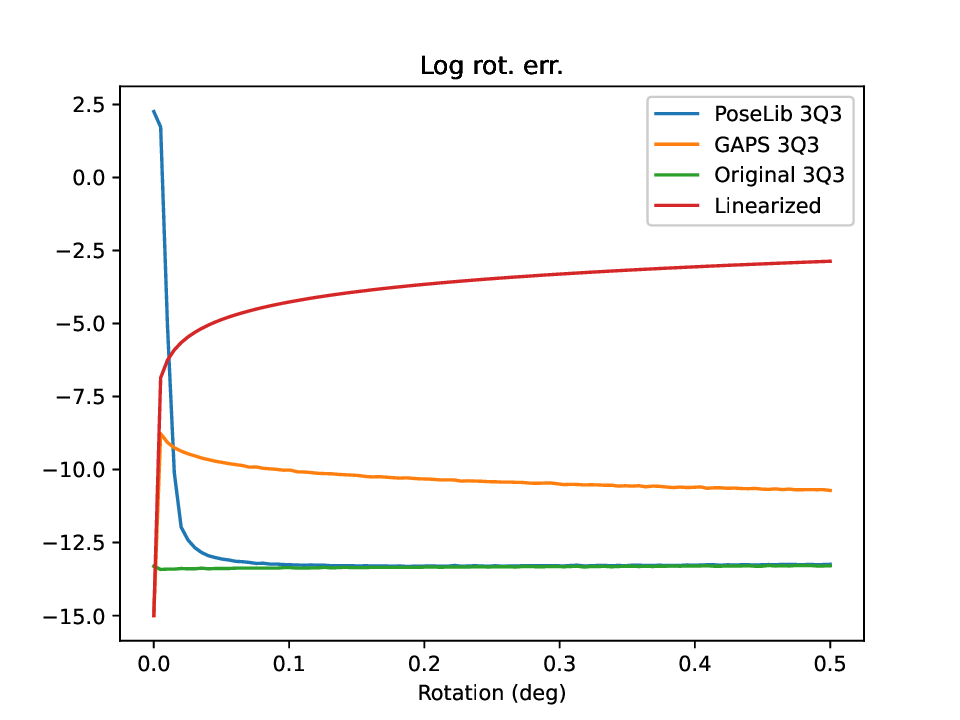}
    \includegraphics[width=0.45\textwidth]{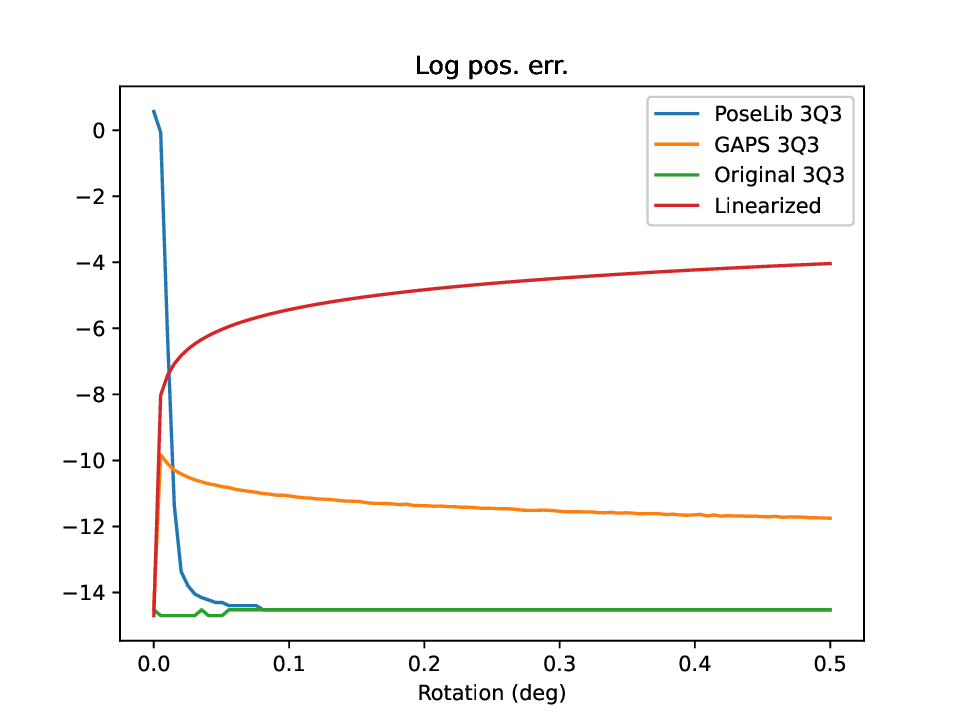}
    \caption{Plot of rotation and position error of various solvers when the ground truth rotation solution has small magnitude ($\leq 0.5^\circ$). The PoseLib 3Q3 implementation is numerically unstable for very small rotations ($< 0.05^\circ$).}
    \label{fig:rotation}
\end{figure}

When the rotation is equal to or close to identity, we found that the performance of our P1AC solver depends on the implementation of 3Q3 solver used.  When using the 3Q3 implementation from PoseLib~\cite{PoseLib}, the solver often produces zero solutions or inaccurate solutions.  This is likely because the last column of the coefficient matrix, after Gaussian elimination, contains extremely small numbers.  When using the original 3Q3 implementation provided by the authors \cite{kukelova2016efficient}, the accuracy remains stable.  

In the specific case where the rotation is identity and the translation is zero, both the PoseLib 3Q3 implementation and the original 3Q3 implementation often fail to produce a solution.  However, we found that a 3Q3 solution produced by the GAPS automatic generator \cite{li2020gaps} avoids all instability with near-identity rotations, even in the case of zero translation.  Using the 3Q3 solver produced by GAPS, the P1AC solver has an average timing of 15 $\mu$s.

We see two possible solutions to the issue encountered with the PoseLib 3Q3 and near-identity rotations, if the P1AC solver is to be used in a scenario where the rotation is expected to be close to identity.  One option is to use the GAPS 3Q3 solver, at the cost of a slightly slower minimal solver.  A faster option is to append the solution from a linearized rotation solver (described below) to the list of PoseLib 3Q3 solutions.  

Figure \ref{fig:rotation} plots the average rotation error for the various solvers over a range of rotation magnitude settings.  For each rotation magnitude setting we generated 10,000 random problems.

\section{Linearized rotation solution}
\label{sec:linearized}

When the rotation is close to identity, we can use a small-angle approximation to arrive at a linear solution to the P1AC problem.  Let $\Vector{r}$ be the $SO(3)$ representation of the rotation $\Matrix{R}$.  We linearize $\Matrix{R}$ using the first order Taylor expansion  $\Matrix{R} \approx \Matrix{I}_{3\times3} + \Skew{r}$. 
 Now the P1AC equations become six linear equation in six unknowns $\Vector{r},\Vector{t}$ and are easily solved.

The solver is extremely fast, with an average timing of 0.366 $\mu$s.

\section{Rotation matrix solution}
\label{sec:nullspace}

The P1AC formulation presented in the main paper is based on the Cayley parameterization of the rotation matrix. As mentioned in the main paper, this parameterization introduces a degeneracy for rotations of exactly 180$^\circ$. Even though this degeneracy is not a problem in practice, we will present here a formulation of the P1AC problem and a solution that does not suffer from this degeneracy.

If we parameterize the rotation directly as a $3\times3$ matrix, then we have 12 parameters in $\Matrix{P}=[~\Matrix{R}~|~\Vector{t}~]$.  We can write the six P1AC equations as a matrix-vector multiply:
\begin{equation}
\Matrix{M}\Matrix{\bar{P}} = \Vector{0}.
\end{equation}
where $\Matrix{\bar{P}}$) denotes the matrix $\Matrix{P}$ rearranged into a vector.

Let $\Matrix{B}$ be the $12\times6$ nullspace of $\Matrix{M}$. We compute the nullspace of $\Matrix{M}$ using singular value decomposition (SVD).
 Any solution, up to scale, for $\Matrix{\bar{P}}$ has the form
\begin{equation}
\label{eq:nullspace}
\Matrix{\bar{P}}=\Matrix{B}\Vector{b},
\end{equation}
where $\Vector{b}$ is a vector of six coefficients for the basis vectors in $\Matrix{B}$.  Our goal is to find solutions for $\Vector{b}$ that make the rotation matrix orthogonal.

We follow the solution procedure described by Ventura et al.~\cite{ventura2014minimal}.  Assuming $b_6 \neq 0$, we remove one parameter and simplify the solution by fixing $b_6=1$.  
Let $\Vector{r}_1, \Vector{r}_2, \Vector{r}_3$ be the rows of $\Matrix{R}$ and $\Vector{c}_1$, $\Vector{c}_2$, $\Vector{c}_3$ be the columns.   The following constraints ensure that $\Matrix{R}$ is orthogonal, up to scale:
\begin{align}
||\Vector{r}_1||^2-||\Vector{r}_2||^2 = 0, \quad
||\Vector{r}_1||^2-||\Vector{r}_3||^2 = 0, \nonumber \\
||\Vector{c}_1||^2-||\Vector{c}_2||^2 = 0, \quad
||\Vector{c}_1||^2-||\Vector{c}_3||^2 = 0, \nonumber \\
\Vector{r}_1 \cdot \Vector{r}_2 = 0, \quad
\Vector{r}_1 \cdot \Vector{r}_3 = 0, \quad
\Vector{r}_2 \cdot \Vector{r}_3 = 0, \nonumber \\
\Vector{c}_1 \cdot \Vector{c}_2 = 0, \quad
\Vector{c}_1 \cdot \Vector{c}_3 = 0, \quad
\Vector{c}_2 \cdot \Vector{c}_3 = 0. \label{eq:10constraints}
\end{align}

Plugging in equation \ref{eq:nullspace} to these constraints results in a system of ten quadratic equations in twenty-one monomials with variables $b_1,\ldots,b_5$.  After extracting the roots of this system of equations, for each solution we divide $\Vector{\tilde{P}}$ by $||\Vector{c_1}||$  and negate $\Vector{\tilde{P}}$ if necessary to ensure that $\textrm{det}(\Matrix{R})=1$.

This system has eight solutions and can be solved using the action matrix method and an automatic solver generator \cite{larsson2017efficient}.
The resulting solver involves elimination of a $47\times55$ template matrix and eigendecomposition of an $8\times8$ matrix. 

The solver is much slower than the 3Q3 solver, having an average timing of 27 $\mu$s.  It does not exhibit any instability with near-identity rotations.

In contrast to \cite{ventura2014minimal}, we discovered that the system of equations \ref{eq:10constraints} can be further simplified by eliminating one unknown, e.g., $b_1$. This can be done by rewriting the ten equations \eqref{eq:10constraints} in a matrix form $\Matrix{C}\Vector{v} = \Vector{0}$, where $\Matrix{C}$ is a $10\times21$ coefficient matrix, and $\Vector{v}$ is a vector of 21 monomials ordered using the lexicographic ordering. After eliminating the matrix $\Matrix{C}$, six monomials containing $b_1$ can be expressed as quadratic polynomials in $b_2,\ldots,b_5$. In this way, $b_1$ can be eliminated from the original equations. Moreover, new equations that express relationships between different monomials can be added to the original equations, e.g. if $b_1 = p_1(b_2,\ldots,b_5)$ and $b_1b_2 = p_2(b_2,\ldots,b_5)$, where $p_1$ and $p_2$ are polynomials in $b_2,\ldots,b_5$,  extracted from the eliminated matrix $\Matrix{C}$, then a new equation that can be added to the original equations has the form $p_1b_2 = p_2$. In this way, a new system of polynomials equations in four unknowns can be generated. This system can be solved using the automatic generator~\cite{larsson2017efficient} and results in a solver that performs elimination of a $29\times 37$ template matrix and eigendecomposition of an $8\times8$ matrix.  Although this solution path is faster than the $47 \times55$ path, we found that it is less numerically stable.

{\small
\bibliographystyle{ieee_fullname}
\bibliography{References}
}